\documentclass[letterpaper]{article}
\usepackage{uai2020}
\usepackage[margin=1in]{geometry}

\usepackage{times}
\usepackage{chicago}
\newcommand{\commentout}[1]{}
\usepackage{graphicx}
\usepackage{amsmath,amsfonts,amsthm,amssymb,mathrsfs} 

\newtheorem{theorem}{Theorem}[section]

\newtheorem{definition}{Definition}[section]

\title{Bounded Rationality in Las Vegas: Probabilistic Finite Automata Play Multi-Armed Bandits}

\author{ {\bf Xinming Liu}\\
Computer Science Dept. \\
Cornell University\\
Ithaca, NY 14853 \\
xl379@cornell.edu \\
\And
{\bf Joseph Y. Halpern}\\
Computer Science Dept. \\
Cornell University 414 Gates Hall\\
Ithaca, NY 14853 \\
halpern@cs.cornell.edu  \\
}

\begin{document}

\maketitle

\begin{abstract}
While traditional economics assumes that humans are fully rational
agents who always maximize their expected utility, in practice, we
constantly observe apparently irrational behavior. One explanation is
that people have limited computational power, so that they are, quite
rationally, making the best decisions they can, given their
computational limitations.  To test this hypothesis, we consider the
multi-armed bandit (MAB) problem. We examine a simple strategy for
playing an MAB
that can be
implemented easily by a probabilistic finite automaton (PFA).
Roughly speaking, the PFA sets certain expectations, and plays an arm
as long as it meets them. If the
PFA has sufficiently many states, it performs near-optimally. Its
performance degrades gracefully as the number of states
decreases. Moreover,
the PFA acts in a ``human-like'' way,
exhibiting a number of standard human biases, like an  \emph{optimism
  bias} and a \emph{negativity bias}.
\end{abstract}

\section{INTRODUCTION}
Behavioral economists have argued for years that the traditional model of \emph{homo economicus}---%
an
agent who is always rational and behaves optimally---is misguided.
 There is a lot of experimental work backing up their claims
(see, e.g., \cite{Thaler15}). Recent work has argued that perhaps the
behavior that we observe can best be explained by thinking of agents
as rational (i.e., trying to behave optimally), but not able to due to
computational limitations; that is, they are doing the best they can,
given their computational limitations. 

In this paper, following a tradition that goes back Rubinstein
\citeyear{rub85} and Neyman \citeyear{ney85}, we model computationally
bounded agents as probabilistic finite automata (PFAs).  We can think
of the number of states of the automaton as a proxy for how
computationally bounded the agent is. Neyman \citeyear{ney85} showed
that cooperation can arise if PFAs play a finitely-repeated prisoner's
dilemma; work on this topic has continued to attract attention (see
Papadimitriou and Yannakakis \citeyear{PY94} and the references
therein). Wilson \citeyear{W02} considered a decision problem where an
agent must decide whether nature is in state 0 or state 1, after
getting signals that are correlated with nature's state.  She
characterized an optimal $n$-state PFA for making this decision, and
showed that it exhibited ``human-like'' behavior; specifically, it
ignored evidence (something a Bayesian would never do), and exhibited
what could be viewed as a first-impression bias and confirmation bias.
Halpern, Pass, and Seeman \citeyear{HPS12} considered a similar
problem in a dynamic setting, where the state of nature could change
(slowly) over time.  Again, they showed that a simple PFA both
performed well and exhibited the kind of behavior humans exhibited in
games studied by Erev, Ert, and Roth \citeyear{EER10}. 

We continue this line of work, and try to understand
the behavior of computationally bounded agents playing a multi-armed
bandit (e.g., playing slot machines in Las Vegas). Our first step in
doing this is to understand the extent to which optimal play can be
approximated by a PFA without worrying about the number of states
used.  There are a number of notions of optimal that we could
consider.  Here we focus on arguably the simplest one: we compare the
expected average payoff of the automaton after it runs for $N$ steps 
to the expected average payoff of always pulling the
optimal arm of the bandit.
We also assume that the possible payoff of each arm is either 1 or 0
(i.e. success or failure), so the expected payoff of an arm is just
the probability of getting a 1.

There are well-known protocols that use
Bayesian methods (e.g., \emph{Thompson Sampling} \cite{Thompson33})
that approach optimal play in the limit; however, these approaches are
computationally expensive.  We show that they have to be.  No approach
that can be implemented by a PFA can perform optimally. Indeed, for
all PFAs, there exists an $\epsilon>0$ such that as the number of
steps gets large, the ratio of the expected payoff of the automaton to
the expected payoff of the optimal arm is at most $1-\epsilon$. That
is, a PFA must be off by some $\epsilon > 0$ from optimal play
(although we  can make $\epsilon$ as small as we like by allowing
sufficiently many states).   Among families of finite automata that
have near-optimal payoff, we are interested in ones that (a) make
efficient use of their states (so, for a fixed number $M$ of states,
have high expected payoff), (b) converge to near-optimal behavior
quickly, and (c) use simple ``human-like'' heuristics. 

A standard approach to dealing with multi-armed bandit problem is one
we call \emph{explore-then-exploit}. We simply test each arm $N$ times
(where $N$ is a parameter), and from then on play the best arm (i.e.,
the one with the highest average reward). If the bandit has $K$ arms,
then we need roughly $O(N^K\log(N)\log(K))$ states, since we need to
keep track of 
the possible tuples of outcomes of the tests, as well as two counters,
one to keep track of which arm is being tested, and the other to keep
track of how many times we have played it.

We can greatly reduce the
number of states by essentially using an elimination tournament.
We first compare arm 1 against arm 2, eliminate the worse arm, run the
winner against arm 3, eliminate the worse arm, run the winner against
arm 4, and so on.  The way we compare arm $i$ and $j$ is straightforward:
we alternate playing $i$ and $j$ and use a counter to keep track of
the relative number of successes of $i$.  If the counter hits an
appropriate threshold $M$ (so that $i$ has had $M$ more successes than
$j$), $i$ is the winner; if the counter hits $-M$, then $j$ is the winner.
To do this, we need ${K  \choose 2}2(2M+1) \sim 2K^2M$ states: we need to keep
track of which arms are being played, which arm is currently moving,
and the counter.  In choosing $M$, we need to balance out the desire
not to mistakenly eliminate a good arm (which is more likely to
happen the smaller that $M$ is) with the desire not to ``waste'' too much
time in finding the right arm (since the payoff while we are doing
that may not be so high, particularly if we are playing two arms whose
success probabilities are equal but not very high).
We deal with this by stopping a comparison after an expected number
$N$ of steps. 
(We implement this by stopping
the comparison
with probability $1/N$, which does not
require any extra states.)
As we shall see, this approach, which we call the
\emph{elimination tournament}, does extremely well. 

The \emph{$\epsilon$-greedy} protocol is a slight variant of this
approach: Again, we test for the first $N$ steps, and then play the
current best arm with probability $1-\epsilon$ and a random arm with
probability $\epsilon$.
But this requires infinitely many states,
since we must keep track of the fraction of successes for all arms to
determine the current best arm.  

Clearly neither approach is optimal.  With positive probability, both
explore-then-exploit and the
elimination-tournament protocols will choose a non-optimal arm; from
then on it is not 
getting the optimal reward.  The $\epsilon$-greedy protocol gets a
non-optimal reward with (roughly) probability $\epsilon$. While we can
make all these approaches arbitrarily close to optimal by choosing the
parameters $N$, $\epsilon$, and $M$ appropriately,
they do not satisfy our
third criterion: they don't seem to be what people are doing.  
The $\epsilon$-greedy approach and explore-then-exploit require
an agent to keep track of large amounts of information, while the
elimination tournament alternates between arms at every
step, which may have nontrivial costs.
(Imagine a gambler in Las Vegas who wants to compare two arms that are
at opposite ends of a large room.  Will he really walk back and forth?)

We instead consider an approach that takes as its starting point
earlier work by Rao 
\citeyear{Rao17}, who considered only two-armed bandits, where,
just as for us,
each arm has a payoff in $\{0,1\}$.
She defined a family of PFAs
that act like ``approximate Bayesians''. More precisely, each arm has
an associated rank that represents a coarse estimate of the arm's
payoff probability.
Rao plays the arms repeatedly (using complicated rules to
determine which arm to play next) in order to estimate 
the success probability of each arm, and then chooses the best arm.



While we use ranks, we use them in a very different way from Rao.  We
take as our inspiration Simon's notion of \emph{satisficing}
\cite{Simon56}.  The idea is that an arm will be accepted if its
success probability is above some threshold.  In the words of
Gigerenzer and Gaissmaier \citeyear{GG15}: ``Set an aspiration
level, search through alternatives sequentially, and stop search as soon
as an alternative is found that satisfies the level.''
(We remark that the importance of the aspiration level goes back to
the 1930s in the psychology literature, and has been studied at length
since then; see, e.g., the highly-cited work of Lewin et al. \citeyear{LDFS44}.)
But how do we determine the aspiration level?  This is a nontrivial
issue.  Selten \citeyear{Selten98} and Simon \citeyear{Simon82}
(both Nobel prize winners) discuss this issue at length.  As
Gigerenzer and Gaissmaier \citeyear{GG15} observe, ``The aspiration
level need not be fixed, but can be dynamically adjusted to feedback.''  
In our setting, it is relatively straightforward: we use an \emph{optimism
  bias} \cite{Sharot11}. We start with a high aspiration level
(success probability) $p$, and run a tournament as above between each arm
$k$ and a ``virtual arm'' that has success probability $p$.  Since
this is a virtual arm, we are essentially comparing the
performance of each arm $k$ to our expectation.  If arm $k$ does not
meet our expectation, then we go to the next arm.  If no arm meets our
expectation, we adjust the aspiration level according to this feedback,
by lowering it.  
This requires $KMm$ states, where $M$, as before, is the counter used
to keep track of the relative performance of the arm being tested and
$m$ is the number of ranks.   We call this the
\emph{aspiration-level} approach. 

We get good performance by taking $m \sim K$, so the aspiration-level
approach uses essentially the same 
number of states as the elimination tournament.  Moreover, as
we show by simulation, its performance approach degrades
gracefully as the number of states decreases. Even with
relatively few states, it compares quite favorably to the
$\epsilon$-greedy approach and to Thompson Sampling, although they
require infinitely many states.   More importantly from our perspective,
the aspiration-level approach is quite human-like.
We have already mentioned how it incorporates satisficing, the
adjustment of expectations according to feedback, and an optimism
bias.  But there is more.  Whereas  the elimination-tournament approach
treats the two arms that it is comparing
symmetrically, the aspiration-level approach does not.  
If the virtual arm wins, it just  
means that we try another arm. Moreover, especially initially, we
expect the virtual arm to win because we start out with a high
aspiration level.  On the other hand, if an actual arm wins, that is
the arm we use from then on.  Thus, we want to be relatively quick to
reject an arm, and slow to accept.  This can be be viewed as a
\emph{negativity bias} \cite{KH72}: negative outcomes have a greater
effect than positive outcomes.  The focus on
recent behavior can be viewed as implementing an \emph{availability
  heuristic} \cite{TK73}: people tend to heavily weight their judgments
  toward more recent or available information.   Finally, a short run of
good luck can have a significant  influence, causing an arm to be
played for a long time (or even played forever, if it is enough to get
it accepted).  People are well-known to label some arms as ``lucky''
and keep playing them long after the evidence has indicated
otherwise.
This can also be viewed as an instance of the \emph{status
  quo bias} \cite{SZ88}: people are much more likely to stick with the
  current state of affairs (provided they think it is reasonably good).  

  \section{MULTI-ARMED BANDITS}
  This section provides the necessary background for the rest of the paper. In particular, we (1) briefly review multi-armed bandits, (2) define the notion of optimality we consider, and (3) prove that a PFA cannot be optimal.

\subsection{THE MULTI-ARMED BANDIT PROBLEM}
The multi-armed bandit (MAB) problem is a standard way of modeling the
tradeoff between exploitation and exploration. An agent has $K$ arms
that she can pull.  Each arm offers a set of possible rewards, each
obtained with some probability.  The agent does not know the
probabilities in advance, but can learn them by playing the arm
sufficiently often. Formally, a \emph{$K$-armed bandit} is a tuple
distribution over rewards for arm \(k\). Let $\mu_k$ be the expected
reward of arm \(i\), for $i=k,\ldots, K$. The best expected reward of
$B$ is denoted  \(\mu^*_B = \max_k\{\mu_k\}\). 

We assume for simplicity in this paper that the possible rewards of an
arm are either 0 or 1.  With this assumption, $\mu_k$ is the
probability of getting a 1 with arm $k$. We can easily modify the
protocol to deal with a finite set of possible rewards, as long as the
set of possible rewards is known in advance. We also assume for now
that the distributions $R_k$ do not vary over time.    

\subsection{OPTIMAL PROTOCOLS FOR MAB PROBLEMS}
We are interested in protocols that play MABs (almost)
optimally. Formally, a \emph{protocol} is a (possibly randomized)
function from history to actions. We focus on one particular simple
notion of optimality here, which informally amounts to approaching the
average reward of the best arm. To make this precise, given a protocol
$P$, let $a_t^{P,B}$ be a random variable that denotes the arm played
by protocol $P$ at the $t$th step. Thus,  $\mu_{a_t^{P,B}}$ is the
expected reward of arm $a_t^{P,B}$. It is easy to see that the
expected cumulative reward of protocol $P$
when run for $N$ steps on MAB $B$  is $Cum(P,B, N) =
\sum_{t=1}^N\mathbb{E}[\mu_{a^{P,B}_t}]$. Since the reward for playing
the optimal arm of MAB $B$ for $N$ steps is $N\mu^*_B$, the
\emph{expected regret} is the difference between the cumulative reward
of $P$ and the optimal reward: $Reg(P,B,N) = N \mu^*_B -
Cum(P,B,N)$. Finally, the \emph{average $N$-step regret} of $P$ on $B$
is $AReg(P,B,N) = Reg(P,B,N)/N$. We say that $P$ is \emph{optimal} if
$\lim_{N \rightarrow \infty} AReg(P,B,N) = 0$ for all MABs $B$. 
  
As we observed in the introduction, neither
explore-then-exploit nor the $\epsilon$-greedy protocol is
optimal in this 
sense.  There are Bayesian approaches that are optimal.  We briefly
discuss one: Thompson Sampling \cite{Thompson33}.  Roughly speaking,
at each step, this protocol computes the probability of each arm being
optimal, given the observations.  It then chooses arm $k$ with a
probability proportional to its current estimate that $k$ is the
optimal arm.  It is not hard to show that, with probability 1, the
probability of a non-optimal arm being chosen goes to 0.  (By way of
contrast, the probability of a non-optimal arm being chosen at any
given step with the $\epsilon$-greedy protocol is a constant: at least
$(K-1)\epsilon/K$, if there are $K$ arms.) 

As shown by Kaufman, Kordan, and Munos \citeyear{KKM12},  Thompson
Sampling is optimal in an even stronger sense than what we have
considered so far.  Taking TS to denote Thompson Sampling, not only do
we have $\lim_{N\rightarrow \infty} Reg(TS,B,N)/N = 0$, but there is a
constant $c^*_B$ (that depends on the MAB $B$, but has been completely
characterized) such that $\lim_{N\rightarrow \infty} Reg(TS,N)/\log(N)
= c^*_B$. Moreover, this is optimal; as shown by Lai and Robbins
\citeyear{LR85}, for all protocols $P$ satisfying a minimal technical
condition, we must have $\lim_{N\rightarrow \infty} G(P,B,N)/\log(N)
\ge c^*_B$. That means that Thompson Sampling approaches optimal
behavior as quickly as possible, and its cumulative regret grows only
logarithmically. We mention this because we will be comparing the
performance of our approach to that of Thompson Sampling later.  

\subsection{PROBABILISTIC FINITE AUTOMATA AND NON-OPTIMALITY}
As we said in the introduction, we are interested in resource-bounded agents playing MABs, and we model resource-boundedness using PFAs. A PFA is just like a deterministic finite automaton, except that the transitions are probabilistic. We also want our automata to produce an output (an arm to pull, or no arm), rather than accepting a language, so, technically, we are looking at what have been called \emph{probabilistic finite automata with output} or \emph{probabilistic transducers}.  (This is also the case for all the earlier papers that considered PFAs playing games or making decisions, such as \cite{HPS12,PY94,rub85,W02}.) Formally, a PFA with output is a tuple \((Q, q_0, \Sigma, O, \gamma, \delta)\), where 
\begin{itemize}
    \item $Q$ is a finite set of \emph{states};
    \item $q_0 \in Q$ is the initial state;
    \item $\Sigma$ is the input alphabet (in our case this will
      consist of the observations ``arm $k$ had reward $j$'' for $j
      \in\{0,1\}$);  
    \item $O$ is the output alphabet (in our case this will be
            ``$k$'', which is interpreted as playing arm $k$, for $k \in\{ 1,
      \ldots, K\}$);  
    \item $\gamma: Q \rightarrow \Delta(O)$ is a probabilistic action function (as usual, $\Delta(X)$ denotes the set of probability distributions on $X$);
    \item $\delta : Q \times \Sigma \rightarrow \Delta(Q)$ is a probabilistic transition function.
\end{itemize} 
Intuitively, the automaton starts in state $q_0$ and plays an arm according to distribution $\gamma(q_0)$.  It then observes the outcome $o$ of pulling the arm (an element of $\Sigma$) and then transitions to a state $q'$ (according to $\delta(q_0,o)$).  It then plays arm $\gamma(q')$, and so on.

It is easy to see that the explore-then-exploit protocol can be
implemented by a finite automaton.  On the other hand, the
$\epsilon$-greedy protocol and Thompson Sampling cannot.  That is
because they keep track of the total number of times each arm $k$ was
played, and the fraction of those times that a reward of 1 was
obtained with $k$.  This requires infinitely many states. 

We claim that no protocol implemented by a PFA can be optimal. To prove this, we need some definitions.
\begin{definition}
A $K$-arm MAB $B = (\mu_1, \ldots, \mu_K)$ is \emph{generic} if (1) $\mu_B^* < 1$, (2) $\mu_i \ne \mu_j$ for $i \ne j$, and (3) if $K=2$, then $\min(\mu_1,\mu_2) > 0$.
\end{definition}
Note the if we put the obvious uniform distribution on the set of $K$-armed bandits (identifying a $K$-armed bandit with a $K$-vector of real numbers), then the set of generic MABs has probability 1.  
\begin{definition}
$B' = (\mu_1', \ldots, \mu_K')$ is a \emph{permutation} of $B =
  (\mu_1, \ldots, \mu_K)$ if there is some permutation $\rho$ of the
    indices such that $\mu_k = \mu'_{\rho(k)}$. 
\end{definition}

\begin{theorem}\label{thm:non-optimal} For all PFAs $M$ and all generic MABs $B$, there exists some $\epsilon_{M,B} > 0$ (that, as the notation suggests, depends on both $M$ and $B$) and an MAB  $B'$ that is a permutation of $B$  such that $\lim_{N\rightarrow \infty} Reg(M,B',N)/N \ge \epsilon_{M,B}$.
\end{theorem}

Before giving the proof, we can explain why we must consider generic
MABs and permutations. To understand why we consider permutations,
suppose that $M$ always plays arm 1.  If it so happens that arm 1 is
the best arm for $B$, then $M$ gets the optimal reward with input $B$.
But it will not get the optimal reward for a permutation of $B$ for
which arm 1 is not the best arm. It is not hard to see that if
$\mu^*_B =1$, then there exists a PFA $M$ that gets the optimal reward
given input $B$ or any of its permutations: $M$ just plays an arm
until it does not get a payoff of 1, then goes on to the next arm.
Sooner or later $M$ will play an arm that always gets a reward of 1. A
similar PFA also gets the optimal reward if $K=2$ given an input $B =
(\mu_1,\mu_2)$ such that  $\mu_k = 0$ for some arm $k$: it alternates
between the arms until it finds an arm that gives reward 1, and sticks
with that arm. Finally, if $\mu_1 = \cdots = \mu_K$, then no matter
what arm $M$ plays, it will get the optimal reward on $B$ and all of
its permutations. The requirement that all $\mu_k$s are distinct is
actually stronger than we need, but since slight perturbations of the
rewards of an arm suffice to make all rewards distinct, we use it here
for simplicity. 

\begin{proof}
Given a PFA $M$ and a nontrivial MAB $B$, there are two possibilities:
(1) there is some state $q$ that can be reached from the start state
$q_0$ with positive probability and an arm $k$ such that, after
reaching state $q$, 
$M$ plays arm $k$ from then on, no matter what it observes; (2)
there is no such state $q$.  Note that the first case is what happens
with explore-then-exploit.  After the exploration phase, the
same arm is played over and over.  The second case is more like
Thompson Sampling or $\epsilon$-greedy; there is always some positive
probability that a given arm $k$ will be played. 

For case (1), let $o_1, \ldots, o_T$ be a sequence  of observations
that, with positive probability, leads $M$ to a state $q$ after which
it always plays arm $k$. If the arm that $M$ plays in state $q$ is not
the best arm of $B$, let $\delta = \mu^*_B - \mu_k$, and let
$\delta_{M,B}$ be the probability with which $o_1, \ldots, o_T$ is
observed when running $M$ on input $B$. Clearly, $\lim_{N\rightarrow
    \infty} Reg(M,B,N)/N \ge \delta \delta_{M,B}$. And if $\mu_k =
\mu^*$, consider a permutation $B' = (\mu_1', \ldots, \mu_K')$ such
that $\mu_j' = 0$ if and only if $\mu_j = 0$ (i.e., the permutation is
the identity on all arms $j$ such that $\mu_j = 0$) such that $\mu_k'
\ne \mu^*_{B'} = \mu^*_B$.  It is still the case that $o_1, \ldots,
o_T$ can be observed with some positive probability $\delta_{M,B'}$
when running $M$ on input $B'$.  Taking $\delta' = \mu^*_B - \mu_j$,
we have $\lim_{N\rightarrow \infty} Reg(M,B,N)/N \ge \delta'
\delta_{M,B'}$. 

For case (2), no matter what state $q$ $M$ is in, with some probability $\epsilon_q > 0$, $M$ plays a non-optimal arm at $q$ or moves to another state $q'$ and plays a non-optimal arm there.  Let $\epsilon_M^* = \min_q \epsilon_q$.  Since $M$ has only finitely many states, $\epsilon^* > 0$. Given as input an MAB $B$, let $\delta_B$ be the difference between the $\mu^*_B$ and the probability that the second-best arm returns 1.  (Here we are using the fact that all arms have different probabilities of returning 1.)  Let $X_{M,B,T}$ be a random variable that represents the reward received on the $T$th step that $M$ is run on input $B$.  Our discussion shows that, for all $T$, we must have $E(X_T + X_{T+1}) \le 2\mu^*_B - \epsilon^*_M\delta_B$, since with probability at least $\epsilon^*_M$, one of $X_T$ or $X_{T+1}$ is at least $\delta_B$ less than $\mu^*_B$. Since $Reg(M,B,2N)$ = $2N\mu^*_B - \sum_{T=1}^{2N} X_{M,B,T} \ge 2N\mu^*_B - N(2\mu^*_B - \epsilon^*_N\delta_B)$, it follows that $Reg(M,B,2N)/2N \ge \epsilon^*_B\delta_B/2$. This gives us the desired result.
\end{proof}

\section{AN ALMOST-OPTIMAL FAMILY OF PFAS FOR MAB PROBLEMS}
In this section, we introduce the aspiration-level protocol more formally.
We start by reviewing
Rao's \citeyear{Rao17} approach to dealing with 2-armed bandits, since
our approach uses some of the same ideas. 

\subsection{RAO'S APPROACH}
With only finitely many states, a PFA cannot keep track of the exact
success rate of each arm in an MAB.  Thus, it needs to keep a finite
representation of the success rate. Rao's idea was to use a finite set
of possible ranks to encode the agent's belief about the relative
goodness of each arm.    There are $m$ possible ranks, $\{1,\ldots,
m\}$, where $m$ is a parameter of the protocol.  Thus, Rao's PFA has
$m^2$ possible states, which have the form $(r_1,r_2)$ (since Rao
considers only 2-armed bandits), where $r_1, r_2 \in \{1,\ldots,
m\}$. 

Rao assumes that the initial state of the PFA has the form $(n,n)$ for some $n \in \{1,\ldots,m\}$; the exact choice does not matter.  Thus, initially, the two arms are assumed to be equally good.  Of course, if an agent has some prior reason to believe that one arm is better than the other, then the initial state can encode this belief.

The action function $\gamma$ is defined as follows: If the
higher-ranked arm has the highest possible rank ($m$) and the other
arm does not, then the higher-ranked arm is played.  Otherwise,
similar in spirit to Thompson Sampling, the next arm to play is chosen
according to a probability that depends on the difference between the
ranks of the arms ($|r_1 - r_2|$) and how far the arm's ranks are from
average ($|r_1 - m/2| + |r_2 - m/2|$).  The two numbers are then
combined using two further parameters (called $\alpha$ and $C_\alpha$
by Rao) of the protocol.  We refer the reader to \cite{Rao17} for the
technical detail and intuition. 

Finally, the transition function $\delta$ is defined as follows: the rank of the arm last played goes up with some probability (if it is not already $m$) if a payoff of 1 is observed and goes down with some probability (if it is not already 1) if a payoff of 0 is observed. The rank of the arm not played does not changed. The exact probability of a state change depends on a quantity that Rao calls the \emph{inertia}, which is determined by the ranks of the arms, and two other parameters of the protocol, called $\beta$ and $C_t$ by Rao. Intuitively, the inertia characterizes the resistance to a change in rank. The less frequently an arm has been played, the higher its associated inertia will be, so its rank is updated with a lower probability. Again, we refer the reader to \cite{Rao17} for details.

\subsection{THE ASPIRATION-LEVEL PROTOCOL}
We want to define a family of PFAs for $K$-armed MABs.  We continue to
use Rao's idea of associating with each arm a rank. The naive
extension would thus require $O(m^K)$ states.  For large $K$, this is
quite unreasonable.
So we assume that the PFA focuses only one arm at a time, comparing it
to a ``virtual'' arm whose success probability can be thought of as
the agent's \emph{aspiration level} \cite{LDFS44}.  The first arm
that meets the agent's aspirations is the arm that is played from then
on.  As we mentioned in the introduction, this can be viewed as
\emph{satisficing} \cite{Simon56}. 
Not only does this approach use significantly fewer states, it seems
more like what people do.  

Rao's protocol has another feature that renders it an implausible
model of human behavior.  It uses a number of parameters ($m, \alpha,
C_\alpha, \beta, C_t$) to trade off exploitation and exploration; the
best choice of parameter settings depends on the application domain.
Moreover, these parameters are combined in a nontrivial way (using,
for example, exponentiation). It is hard to believe that people would
take the trouble (or have enough experience) to learn the appropriate
parameter settings for a particular domain, nor are they likely to be
willing to do the computations needed to use them. 

We thus significantly simplify the action function and transition
function.  
As we said, we use the idea of a tournament, but we play the current
arm against a ``virtual arm'', whose success probability is determined
by the aspiration level, which is rank.  If there are $m$ ranks, then
a rank of $r \in \{1, \ldots, m\}$ can be thought of as representing
the interval of probability $[(r-1)/m, r/m]$.  We thus take the
success probability of a virtual arm with aspiration level $r$ to be
$(r-.5)/m$, the midpoint of the interval.  If we compare arm $i$ to
the virtual arm using a counter.  Suppose that we get a success with
arm $i$ (i.e., 1 is observed).  Since we expect the virtual arm to
have a success with 
probability $(r-.5)/m$, we increase the counter by 1 with probability
$1-(r-.5)/m$ (since this is the probability that the virtual arm had a
failure, so that arm $i$ had one more success than the virtual arm),
and leave the counter unchanged with probability $(r-.5)/m$ (since,
with this probability, both the virtual arm and arm $i$ had a
success).  Similarly, if there is a failure with arm $i$, we decrease
the counter with probability $(r-.5)/m$ and leave it unchanged with
probability $1-(r-.5)/m$.

We use two thresholds $M_1$ and $M_2$ to decide when to end the
comparison.  If the counter reaches $M_1$, then we declare the current
arm $i$ being considered to
have won the tournament; intuitively, its success probability is
higher than that 
of the virtual arm.  From then on we play arm $i$.  If the counter
reaches $M_2$, then the virtual arm has won the comparison.  We
(temporarily) eliminate arm $k$, and compare the virtual arm to arm
$k+1$ if $k < K$.  We discuss what happens if $k=K$ shortly, but first note
that there is no analogue to the parameter $N$ of the 
elimination-tournament protocol here.  The concern in the
elimination-tournament protocol is that we are comparing two arms $i$
and $j$ that 
have roughly equal, but not very good success probabilities.  Then the
tournament will go on for a long time, but not give a high reward.
With the aspiration-level protocol, if arm $k$ has a success
probability that is essentially 
the same as that of the virtual arm, although the comparison may go on
for a long time, the agent is getting a cumulative reward that
essentially matches expectations, so there is no pressure to stop the
comparison.

If $k=K$, then the virtual arm did better than all arms with this
aspiration level.  That means that our expectations are too high, so we lower
the aspiration level from $r$ to $r-1$, and retest all arms.

As discussed in the introduction, we do not assume that $M_1 = M_2$.  
The implications of an arm $i$ winning the comparison against the virtual
arm are much different than the implications of the virtual arm
winning.  In the former case, we play arm $i$ from then on; in the
latter case, we just continue looking for another (hopefully better)
arm.  Because the implications are so different, it turns out that we
want to take $M_1$ significantly larger than $M_2$.  (Our experiments
suggest that $M_1 = 20$ and $M_2 =-3$ are good choices, along with
$m=100$; see Section~\ref{sec:apirationparameters}.)

One other issue: if the actual best success probability is low
(say, .2) and there are 100 ranks, it will take a long time before the
aspiration level is set appropriately.  During this time, the
cumulative regret is increasing.  To speed up the process of finding
the ``right'' aspiration level, we can do a quick preprocessing phase to
find the right range, and then explore more carefully.  Specifically,
if $m=100$, in the preprocessing phase, when we reset the rank, we
decrease it by 10 (in general, we decrease it by $\sqrt{m}$) rather
than decreasing it by 1.  We also use smaller values of $M_1$ and
$M_2$ (say, $M_1=5$ and $M_2 = -1$ rather than $M_1 = 20$ and $M_2=
-3$).  If an arm $i$ beats the virtual arm when the aspiration level
$r= 60$, we go back to the previous setting of aspiration level
$r=70$, and do a more careful search starting from there, now
decreasing the aspiration level by 1, and using $M_1 = 20$ and
$M_2=-3$.  This preprocessing phase allows us to home in on the
appropriate expectations quickly.   Again, besides being more
efficient, this seems to be the type of thing that people do.

With this background, we are ready to define our family of
PFAs.
For ease of presentation, we do not use a preprocessing phase.  
Formally, we have a family $M_{K, m, M_1,M_2} =
(Q_{K,m,M_1,M_2}, q_m,\Sigma_K, O_K, \gamma_{K},
\delta_{K,m,M_1,M_2})$  of PFAs, indexed by 4 parameters: \(K\) is the
total number of arms,
\(m\) is the number of possible ranks for each arm, and $M_1$ and
$M_2$ are the upper and lower thresholds for the counter.
We assume that $K$ is given as part of
the input; we discuss how $m$, $M_1$, and $M_2$ are chosen in the
next section.  Not only do we have fewer parameters than Rao, as we
shall see, they are easier to set (and easier to explain and
understand). In more detail, the components of the tuple are as
follows: 
\begin{itemize}
  \item A state $q \in Q_{K,m,M_1,M_2}$ has the form $(r,k, c)$,
    where $1 \le r \le m$, $1 \le k \le K$, and $-M_2 < c \le M_1$.
    Intuitively, a state $(r,k,c)$ says that the current aspiration level
    is $r$, we are testing arm $k$, and the counter that keeps track of
    the relative success rate of arm $k$ compared to the virtual arm is at $c$.

\item We take the initial state $q_0$ to be $(m, 1, 0)$: 
we start by setting the aspiration level to $m$ (the highest level
possible), testing arm 1, and have the counter at 0.

\item $\Sigma_K$ consists of observations of the form $(k,h)$, where
  $k \in \{1, \ldots, K\}$ and $h \in \{0,1\}$.  We observe the
  outcome of playing arm $k$, which is a reward of either 0
  or 1. 
  
  \item $O_K = \{1, \ldots, K\}$: we can play any arm.
  
  \item The action function $\gamma_{K}$ at a state $(r,k,c)$ 
plays arm $k$.
  
  \item The transition function $\delta_{K,m,M_1,M_2}$ proceeds as
        follows.  In state $(r,k,c)$,
if $c=M_1$, the state does not change.  (We have chosen $i$ as the arm
to play from then on.)  If $c < M_1$, given an observation $(h,k)$, if 
$h=1$ (a success was observed), the new state is $(r,i,c')$, where
$c'=c+1$ with 
probability $1-(r-.5)/m$, and otherwise $c' = c$.  If $h=0$ and $c>
M_2+1$, then the new state is $(r,i,c')$, where $c'=c-1$ with
probability $(r-.5)/m$, and otherwise is unchanged.  If $c = M_2+1$,
then with probability $(r-.5)/m$, the new state is $(r-1,1,0)$ (the
aspiration level is lowered and we start over comparing the virtual
arm to all the arms, starting with arm 1); otherwise the state is unchanged.
\end{itemize}

\section{EXPERIMENTS}
\subsection{PERFORMANCE METRICS}
We use simulations to test the performance of various protocols. In
the simulations, we consider an MAB $B$ with $K$ arms, whose
true success probabilities are uniformly distributed in [0,$\alpha$],
where $\alpha$ is a random number in [0,1]. If we had just assumed
that the success probabilities were uniformly distributed in [0,1], 
then the probability of there being 
an arm in the [0.9,1] interval is $1-0.9^K$, which is approximately
0.995 for $K=50$.  Indeed, the probability of there being an arm in the
interval $[.99,1]$, is about 0.4.  Not only does this seem
unreasonable in practice, this assumption would make it too easy to
set the right aspiration level in our approach (i.e., it would hide
some real-world difficulty).
The assumption that the success probabilities are bounded by $\alpha$ for a
randomly-chosen $\alpha$ seems more reasonable.  While assuming that
the success probabilities are \emph{uniformly} distributed in
$[0,\alpha]$ may not be so reasonable, our results remain essentially
unchanged even if the success probabilities are chosen adversarially,
and the uniform distribution is much easier to generate.

We focus on two metrics when it comes to measuring the performance of a
protocol: 
(1) the expected cumulative regret of a protocol $P$ as a function of the
number of steps played  (which roughly depends on how long it
takes to find the best arm) and (2) the expected average regret in the limit
(i.e., $\lim_{N \rightarrow \infty} AReg(P,B,N)$), which essentially measures the gap
between the success probability of the arm chosen by protocol $P$ and
the success probability of the optimal arm of $B$.  We take the 
expectation over MABs $B$ generated as discussed above.  Essentially,
we want a protocol that gets to the best arm quickly and
accurately.

\subsection{PARAMETER SETTINGS IN THE ASPIRATION-LEVEL PROTOCOL}\label{sec:apirationparameters}
There are three parameter settings for the aspiration-level
protocol: the number of ranks $m$, and the thresholds $M_1$ and $M_2$
for winning and losing a comparison against the virtual arm.
We examine the effect of different choices here.

The larger $m$ is, the finer distinctions we will be able to make
between the arms that we are testing.  Roughly speaking, if the
virtual arm has rank $r$ and the virtual arm performed better than all
arms when the aspiration level was $r+1$, we would expect that all
arms have success probability less than $(r+.5)/m$, and that an arm
with success probability greater than $(r-.5)/m$ will beat the virtual
arm.  However, this arm can have probability as much as $1/m$ less than
the arm with highest success probability.  By taking $m$ larger, we
thus minimize the expected gap between the success probability of the
arm chosen and the best arm.  

\begin{figure}[h]
    \includegraphics[width=6.5cm]{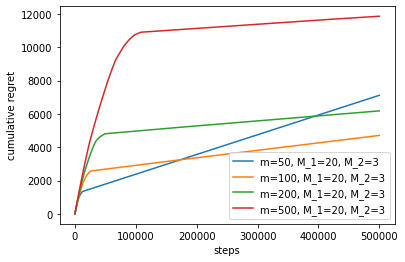}
    \caption{Cumulative regret for different $m$.}
    \label{fig:1}
\end{figure}

We consider an MAB $B$ with $K=50$ arms and run simulations. As
expected, the larger $m$ is, the smaller the gap, but there are
diminishing returns. Figure \ref{fig:1} shows that, with other
parameters fixed ($M_1=20, M_2=3$), there is significant improvement in
going from $m=50$ to $m=100$; but the marginal improvement drops off
quickly. This is no significant difference between $m=100$ and larger
values such as $m=200$ or $m=500$. The corresponding gaps between the
success probability of the arm chosen and the success
probability of the optimal arm of $B$, averaged over 100 repetitions,
are 0.020, 0.007, 0.0068, 0.0065, respectively. In addition, since we
start optimistically by initializing the aspiration level at the
highest possible rank, when $m$ is larger, it takes longer to get the
right aspiration level and hence the cumulative regret is larger, as
shown in Figure \ref{fig:1}. Considering both performance metrics as
mentioned above, we choose $m=100$ in the later simulations. 

\begin{figure}[h]
    \includegraphics[width=6.5cm]{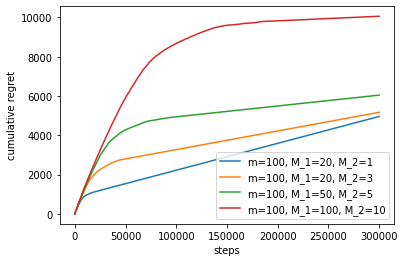}
    \caption{Cumulative regret for different $M_1$ and $M_2$.}
    \label{fig:2}
\end{figure}

Once we fix $m$, we now examine the choices of $M_1$ and $M_2$.
The parameters $M_1$ and $M_2$ determine the conditions of winning and
losing: if counter gets to $M_1$, then the current arm beats the
``virtual arm" and is therefore chosen as the best arm; if the counter
gets to $-M_2$, then the current arm loses the tournament with the
``virtual arm" and we move to a new arm. If all $K$ arms lose the
tournament, we decrease the aspiration level by 1 and restart
the tournament. We want it to be easier for the ``virtual arm" to win,
since the consequences are lower in that case (the protocol ends if
we declare arm $i$ a winner, whereas we keep going if the ``virtual
arm" is a winner). Therefore, it makes sense to have an asymmetry and
choose $M_1$ greater than $M_2$.
We again consider an MAB $B$ with
$K=50$ arms and $m=100$ fixed. As shown in Figure \ref{fig:2}, the
cumulative regret increases as $M_1$ and $M_2$ get larger. However,
the gap between the success probability of the arm chosen by protocol
$P$ and the success probability of the optimal arm of $B$,
decreases. The corresponding gaps, averaged over 100 repetitions, are
0.014, 0.007, 0.005, 0.004, respectively. Since the number  of states
in the aspiration-level protocol is $Km(M_1 + M_2)$, there is a
tradeoff between accuracy and the number of states required. Taking
into account state-efficiency, accuracy, and the expected cumulative
regret, we choose $M_1=20$ and $M_2=3$.

Both Figure \ref{fig:1} and Figure \ref{fig:2} show that the
performance of the aspiration-level protocol degrades quite
gracefully as we take 
smaller values of $m$, $M_1$, and $M_2$ (which is how we would have to deal 
with having
fewer states). 

\subsection{PARAMETER SETTINGS FOR THE ELIMINATION TOURNAMENT}
\begin{figure}[h]
    \includegraphics[width=6.5cm]{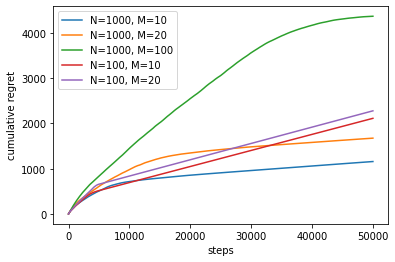}
    \caption{Cumulative regret for different $N$ and $M$.}
    \label{fig:3}
\end{figure}

The elimination-tournament protocol has two parameters: $M$ (the point
at which an arm is declared a winner in the two-way comparison) and
$N$ (recall that $1/N$ is the probability that an arm is declared in the
two-way comparison if no arm is dominant and has $M$ more successes than
the other).    Thus, after an expected number of at most $N(K-1)$
steps, the elimination-tournament protocol has reduced to one arm.  We clearly 
want $M$ and $N$ to be large enough to give the protocol time to select a
relatively good arm. However, we don't want to stick with bad arms for
too long, since this will lead to larger cumulative regret. 
Figure \ref{fig:3} shows the cumulative regrets for different choices
of $N$ and $M$, for an MAB with $K=50$ arms.
For the choices of $(N,M)$ considered---(1000,10), (1000,20), (1000,
100), (100,10), (100,20)---the gaps, averaged 100 repetitions,
are 0.01, 0.007, 0.006, 0.03, 0.03,
respectively. Both $N=1000, M=20$ and $N=1000, M=100$
give similarly good performance in terms of the expected average
regret, but the latter leads to larger expected cumulative
regret. Therefore, for $K=50$, we choose $N=1000$ and $M=20$. 

\subsection{COMPARING PROTOCOLS}
Based on the simulations above, to minimize the number of states used
while maintaining relatively good performance, for $K=50$, we choose the
parameters $m=100, M_1 = 20, M_2 = 3$ for the aspiration-level
protocol and $M=20$, $N=1000$ for the elimination-tournament protocol, and
compare these two finite-state protocols to the 
$\epsilon$-greedy protocol and Thompson Sampling, which are
infinite-state protocols.  With these choices, the
aspiration-level protocol uses 115,000 states, while the
elimination-tournament protocol uses just over 100,000.  While this
may seem to be a a lot of states, they can be encoded using 17 bits.
Given the number of neurons in a human brain, this should not be a problem.

We can greatly reduce the cumulative regret for the
aspiration-level protocol by a preprocessing phase, as
suggested earlier. For $K=50$ arms and the aspiration-level
protocol with $m=100, M_1 = 20, M_2 = 3$, we first use a preprocessing
phase to get a rough idea of what the true highest success probability
might be.
We use the parameters suggested earlier, decreasing the aspiration
level by 10 after testing all the arms in the preprocessing, and use
thresholds $M_1' = -5$ and $M_2'= -1$.  
We use this two-phase approach for the aspiration-level protocol in
the following simulation. 

We consider MABs with $K=50$ arms, and see how
the elimination-tournament protocol, the aspiration-level protocol,
$\epsilon$-greedy, and Thompson sampling perform.
Not surprisingly, Thompson sampling
performs best, and has logarithmic cumulative regret, whereas the
other three protocols have linear cumulative regret.
After 50,000 steps, the
expected difference between the success probability of the arm chosen
and that of the optimal arm for
these protocols are 0.007, 0.008, 0.025, 0.003, respectively.
Interestingly, both the aspiration-level protocol and the
elimination-tournament protocol eventually outperform
$\epsilon$-greedy, although the latter requires
infinitely many states.

\begin{figure}[h]
    \includegraphics[width=6.5cm]{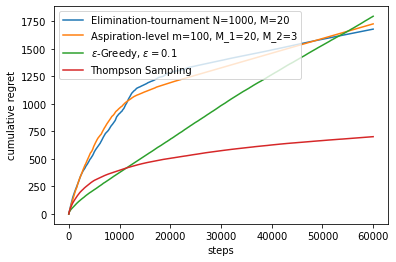}
    \caption{Cumulative regret over time.}
    \label{fig:4}
\end{figure}

\section{DISCUSSION}
We have introduced two finite-state protocols for playing MABs, the
aspiration-level protocol and the elimination-tournament protocol.
Both perform quite well in practice, while using relatively few states.
In cases where switching between arms incurs a significant
cost, the aspiration-level protocol is a better choice. 

Recall that the main motivation for this study was understanding human
behavior.
\commentout{
As we observed, building in a \emph{status quo} bias
improves the performance of our protocol.  It also demonstrates other
``human-like'' behavior. For example, we can think of the focus set
(which we introduced in order to decrease the size of the state space)
as an instance of an availability bias \cite{TK73}: the only arms that
are played are those that are ``available''; that is, those in the
focus set. Finally, a run of bad luck with a particular arm (which can
be viewed as a negative outcome for that arm) can result in result in
a good arm being eliminated.  This is more likely to happen if $k=2$
than if $k$ is large. With smaller values of $m$, we also get more
variability in the ranks, so again, a run of bad luck can have a
larger impact. 
}
The fact that the aspiration-level protocol exhibits such human-like
behavior, including adjusting aspiration levels according to feedback,
an optimism bias, a negativity bias, and a status quo bias, as well as
a focus on recent behavior, suggests that humans are not being so
irrational.
Note that these biases are emphasized if the number of states is
decreased.  For example, if an agent decreases $M_2$, the threshold for rejecting
an arm in a two-way comparison with the virtual arm, in response to
having fewer states, this increases the negativity
bias. Decreasing $M_1$ increases the likelihood that an agent will continue
to play an apparently  ``lucky arm''.  The impact of decreasing $M$
in the elimination-tournament protocol is similar.  
The bottom line is that these protocols exhibit 
apparently irrational behavior for quite rational reasons!  At the
same time, they may be of interest even for those not interested in
modeling human behavior, since they have quite good performance, even
with relatively few states.  

We have focused here on a static setting, where the probabilities do
not change over time.  We could easily modify our PFA to deal with the
dynamic setting by simply
resetting the tournaments from time to time.
More interestingly, we would like to apply these ideas to
a more game-theoretic setting, such as the wildlife poaching setting
considered by Kar et al.~\citeyear{Kar15}, where rangers are trying to
protect rhinos from poachers.  We hope to report on that in future
work. 


\subsubsection*{Acknowledgements}
This research was supported by MURI (MultiUniversity Research
Initiative) under grant 
W911NF-19-1-0217, by the ARO under grant W911NF-17-1-0592,
by  the NSF under grants IIS-1703846 and IIS-1718108, and by a grant
from the Open Philosophy Foundation. We thank Alice
Chen for her preliminary work and comments on an earlier version of
the manuscript. We also thank four anonymous reviewers for their
feedback. 

\eject
\bibliographystyle{chicago}
\bibliography{joe,z}

\begin{thebibliography}{}

\bibitem[\protect\citeauthoryear{Erev, Ert, and Roth}{Erev
  et~al.}{2010}]{EER10}
Erev, I., E.~Ert, and A.~E. Roth (2010).
\newblock A choice prediction competition for market entry games: An
  introduction.
\newblock {\em Games and Economic Behavior\/}~{\em 1\/}(1), 117--136.

\bibitem[\protect\citeauthoryear{Gigerenzer and Gaissmaier}{Gigerenzer and
  Gaissmaier}{2015}]{GG15}
Gigerenzer, G. and W.~Gaissmaier (2015).
\newblock Decision making: Nonrational theories.
\newblock In J.~D. Wright (Ed.), {\em International Encyclopedia of the Social
  and Behavioral Sciences (2nd Edition)}, pp.\  911--916.

\bibitem[\protect\citeauthoryear{Halpern, Pass, and Seeman}{Halpern
  et~al.}{2012}]{HPS12}
Halpern, J.~Y., R.~Pass, and L.~Seeman (2012).
\newblock I'm doing as well as {I} can: modeling people as rational finite
  automata.
\newblock In {\em Proc.~Twenty-Sixth National Conference on Artificial
  Intelligence (AAAI '12)}, pp.\  1917--1923.

\bibitem[\protect\citeauthoryear{Kanouse and Hanson}{Kanouse and
  Hanson}{1972}]{KH72}
Kanouse, D.~E. and L.~Hanson (1972).
\newblock Negativity in evaluations.
\newblock In E.~E. Jones, D.~E. Kanouse, S.~Valins, H.~H. Kelley, R.~E.
  Nisbett, and B.~Weiner (Eds.), {\em Attribution: Perceiving the Causes of
  Behavior}. Morristown, NJ: General Learning Press.

\bibitem[\protect\citeauthoryear{Kar, Fang, {Delle Fave}, Sintov, and
  Tambe}{Kar et~al.}{2015}]{Kar15}
Kar, D., F.~Fang, F.~{Delle Fave}, N.~Sintov, and M.~Tambe (2015).
\newblock ``{G}ame of thrones'': when human behavior models compete in repeated
  {S}tackelberg security games.
\newblock In {\em Proc.~2015 International Conference on Autonomous Agents and
  Multiagent Systems}, pp.\  1381--1390.

\bibitem[\protect\citeauthoryear{Kaufmann, Korda, and Munos}{Kaufmann
  et~al.}{2012}]{KKM12}
Kaufmann, E., N.~Korda, and R.~Munos (2012).
\newblock Thompson sampling: an asymptotically optimal finite-time analysis.
\newblock In N.~H. Bshouty, G.~Stoltz, N.~Vayatis, and T.~Zeugmann (Eds.), {\em
  Algorithmic Learning Theory )ALT 2012)}, LNCS, Volume~7568, pp.\  199--213.
  Springer.

\bibitem[\protect\citeauthoryear{Lai and Robbins}{Lai and Robbins}{1985}]{LR85}
Lai, T.~L. and H.~Robbins (1985).
\newblock Asymptotically efficient adaptive allocation rules.
\newblock {\em Advances in Applied Mathematics\/}~{\em 6\/}(1), 4--22.

\bibitem[\protect\citeauthoryear{Lewin, Dembo, Festinger, and Sears}{Lewin
  et~al.}{1944}]{LDFS44}
Lewin, K., T.~Dembo, L.~Festinger, and P.~S. Sears (1944).
\newblock Level of aspiration.
\newblock In J.~M. Hunt (Ed.), {\em Personality and the Behavior Disorders},
  pp.\  333–378. Cambridge, MA: Ronald Press.

\bibitem[\protect\citeauthoryear{Neyman}{Neyman}{1985}]{ney85}
Neyman, A. (1985).
\newblock Bounded complexity justifies cooperation in finitely repeated
  prisoner's dilemma.
\newblock {\em Economic Letters\/}~{\em 19}, 227--229.

\bibitem[\protect\citeauthoryear{Papadimitriou and Yannakakis}{Papadimitriou
  and Yannakakis}{1994}]{PY94}
Papadimitriou, C.~H. and M.~Yannakakis (1994).
\newblock On complexity as bounded rationality.
\newblock In {\em Proc.~26th ACM Symposium on Theory of Computing}, pp.\
  726--733.

\bibitem[\protect\citeauthoryear{Rao}{Rao}{2017}]{Rao17}
Rao, A. (2017).
\newblock A finite memory automaton for two-armed {B}ernoulli bandit problems.
\newblock In {\em Proc.~Thirty-First National Conference on Artificial
  Intelligence (AAAI '17)}, pp.\  4981--4982.
\newblock The full paper is available at
  http://raoariel.github.io/raoariel-fma.pdf.

\bibitem[\protect\citeauthoryear{Rubinstein}{Rubinstein}{1986}]{rub85}
Rubinstein, A. (1986).
\newblock Finite automata play the repeated prisoner's dilemma.
\newblock {\em Journal of Economic Theory\/}~{\em 39}, 83--96.

\bibitem[\protect\citeauthoryear{Samuelson and Zeckhauser}{Samuelson and
  Zeckhauser}{1998}]{SZ88}
Samuelson, W. and R.~Zeckhauser (1998).
\newblock Status quo bias in decision making.
\newblock {\em Journal of Risk and Uncertainty\/}~{\em 1}, 7--59.

\bibitem[\protect\citeauthoryear{Selten}{Selten}{1998}]{Selten98}
Selten, R. (1998).
\newblock Aspiration adaptation theory.
\newblock {\em Journal of Mathematical Psychology\/}~{\em 42}, 191–214.

\bibitem[\protect\citeauthoryear{Sharot}{Sharot}{2011}]{Sharot11}
Sharot, T. (2011).
\newblock {\em The Optimism Bias: A Tour of the Irrationally Positive Brain}.
\newblock New York, NY: Pantheon Books.

\bibitem[\protect\citeauthoryear{Simon}{Simon}{1956}]{Simon56}
Simon, H.~A. (1956).
\newblock Rational choice and the structure of the environment.
\newblock {\em Psychological Review\/}~{\em 63\/}(2), 129--138.

\bibitem[\protect\citeauthoryear{Simon}{Simon}{1982}]{Simon82}
Simon, H.~A. (1982).
\newblock {\em Models of bounded rationality}.
\newblock Cambridge, MA: MIT Press.

\bibitem[\protect\citeauthoryear{Thaler}{Thaler}{2015}]{Thaler15}
Thaler, R. (2015).
\newblock {\em Misbehaving: The Making of Behavioral Economics}.
\newblock New York, NY: W. W. Norton and Company.

\bibitem[\protect\citeauthoryear{Thompson}{Thompson}{1933}]{Thompson33}
Thompson, W.~R. (1933).
\newblock On the likelihood that one unknown probability exceeds another in
  view of the evidence of two samples.
\newblock {\em Biometrika\/}~{\em 25\/}(3--4), 285--294.

\bibitem[\protect\citeauthoryear{Tversky and Kahneman}{Tversky and
  Kahneman}{1973}]{TK73}
Tversky, A. and D.~Kahneman (1973).
\newblock Availability: a heuristic for judging frequency and probability.
\newblock {\em Cognitive Psychology\/}~{\em 5}, 207--232.

\bibitem[\protect\citeauthoryear{Wilson}{Wilson}{2015}]{W02}
Wilson, A. (2015).
\newblock Bounded memory and biases in information processing.
\newblock {\em Econometrica\/}~{\em 82\/}(6), 2257--2294.

\end{thebibliography}

\end{document}